\documentclass[runningheads]{llncs}
\usepackage{graphicx}
\usepackage{amsmath,amssymb} 
\usepackage{color}
\usepackage{caption}
\usepackage{adjustbox}
\usepackage{multirow}
\usepackage{subfigure}
\usepackage{color}
\usepackage{caption} 

\usepackage[width=122mm,left=12mm,paperwidth=146mm,height=193mm,top=12mm,paperheight=217mm]{geometry}
\begin{document}
\mainmatter

\title{Temporal HeartNet: Towards Human-Level  Automatic Analysis of Fetal Cardiac Screening Video} 

\titlerunning{Temporal HeartNet: Automatic Analysis of Fetal Cardiac Screening Video}

\authorrunning{W. Huang, C. P. Bridge, J. A. Noble, and A. Zisserman}


\author{Weilin Huang, Christopher P. Bridge, J. Alison Noble, and Andrew Zisserman}

\institute{Department of Engineering Science, University of Oxford\\
\email{whuang@robots.ox.ac.uk}
}



\maketitle

\begin{abstract}
We present an automatic method to describe clinically useful information about scanning, and to guide image interpretation in ultrasound (US) videos of the fetal heart. Our method is able to jointly predict the visibility, viewing plane, location and orientation of the fetal heart at the frame level. The contributions of the paper are three-fold: (i) a convolutional neural network architecture is developed for a multi-task prediction, which is computed by sliding a $3 \times 3$ window \textit{spatially} through  convolutional maps.  (ii) an anchor mechanism and Intersection over Union (IoU) loss are applied for improving localization accuracy. (iii) a recurrent architecture is designed to recursively compute regional convolutional features \textit{temporally} over sequential frames, allowing each prediction to be conditioned on the whole video. This results in a spatial-temporal model that precisely describes detailed heart parameters in challenging US videos.
We report results on a real-world clinical dataset, where our method achieves performance on par with expert annotations.

\end{abstract}

\section{Introduction}

Understanding fetal cardiac screening ultrasound (US) videos is crucial to diagnosis of congenital heart disease (CHD). However, because imaging the fetal heart requires expertise, in many countries fetal cardiac screening is not included as a compulsory part of the 20-week abnormality scan, leading to a high mis-diagnosis rate of fetal heart conditions. Detection rates of CHD could be potentially improved through efficient automated analysis to support sonographers as they scan. However, standardization of  ultrasound imaging of the fetal heart acquisition is difficult, and leads to quite varied appearance of the fetal heart in standard planes, as illustrated in Fig. \ref{fig:main}. This varied appearance has meant that it has proven difficult to automate analysis of fetal cardiac screening images. Fetal  cardiac screening videos could be analyzed in a frame-by-frame manner, but videos contain rich spatio-temporal acoustic patterns, the temporal component of which is ignored in a frame-by-frame approach.  The  approach we describe in this paper is to our knowledge the first to develop a deep recurrent convolutional model for this challenging task. In our approach we directly compute a dynamic fetal heart description from a fetal cardiac screening video, including an identification of the standard view. The most closely related work is that of Bridge \textit{et. al.} \cite{Bridge2017} that inspired the current work, but estimates a fetal heart description using a Bayesian (state-space) approach which includes an explicit temporal filtering step. The current paper describes a method that goes from video to fetal heart description and standard viewing plane identification in one step. 

Specifically, we develop a temporal CNN for fetal heart analysis in cardiac screening videos. It jointly predicts  multiple key variables relating to the fetal heart, including the visibility, viewing plane, location and orientation of the heart. Our model is a recurrent convolutional architecture by design. It computes deep features from US images with a CNN, and a RNN is designed to propagate temporal information through frames. We include a number of technical developments that allow a general CNN to work efficiently for our task, with  human-level performance achieved. Our contributions are described as follows:

\begin{itemize}
\item[--] We formulate the problem of fetal heart analysis as a multi-task prediction within a CNN detection framework.  
This allows heart details to be predicted at each local region of image, which is crucial to achieving an accurate estimation of multiple heart parameters.


\item[--] We design circular anchors to handle the approximately circular shape of the heart at multiple scales. Detection accruacy is futher improved by adopting an Intersection over Union (IoU) loss \cite{Yu2016}, which jointly regresses the center and radius of the heart  as a whole unit.

\item[--] We develop a bi-directional RNN that recursively computes local heart details in both temporal directions, allowing it to model heart dynamics over the whole video.  This is of particular importance for describing 2D US video,  where individual image-level features are relatively weak, and the key objects of interest (e.g., a heart valve)  can go in and out of the imaging plane. 
\end{itemize}




\subsection{Related Work}

Automatic methods for US video analysis have been developed. For example, Kwitt \textit{et. al.} \cite{Kwitt2013} applied kernel dynamic texture models for labelling individual frames in an US video.  This concept was extended to handle multiple object classes on real-world clinical US videos in \cite{Maraci2017}. Recently, CNNs have been applied to image-level classification of  anatomical structures with transfer learning \cite{Gao2016} and recurrent models \cite{Chen2015}. These works are all related to frame-level classification of US videos.  Our application differs from them by focusing on describing details of the fetal heart, which is a more complicated multi-task application.

The most closely related work to this paper, is that of Bridge \textit{et. al.} \cite{Bridge2017}, where a number of key parameters related to the fetal heart are estimated. Hand-crafted features were used with classification forests to distinguish different view planes of the fetal heart. Then a multi-variable prediction was formulated that used a CRF-filter. We adress a similar task, but propose a significantly different approach that builds on recent advances in deep learning. Firstly, this allows it to leverage deep, learned feature representations that are shared by all tasks, rather than relying on hand-crafted features.  Secondly, by including a recurrent part, it is possible to train our model end-to-end, 
whereas in \cite{Bridge2017} the classification and regression parts are trained separately from the temporal filter.


\section{Temporal Fetal Heart Network}
\begin{figure}[tb]
\centering
\includegraphics[height=5.5cm, width=11cm]{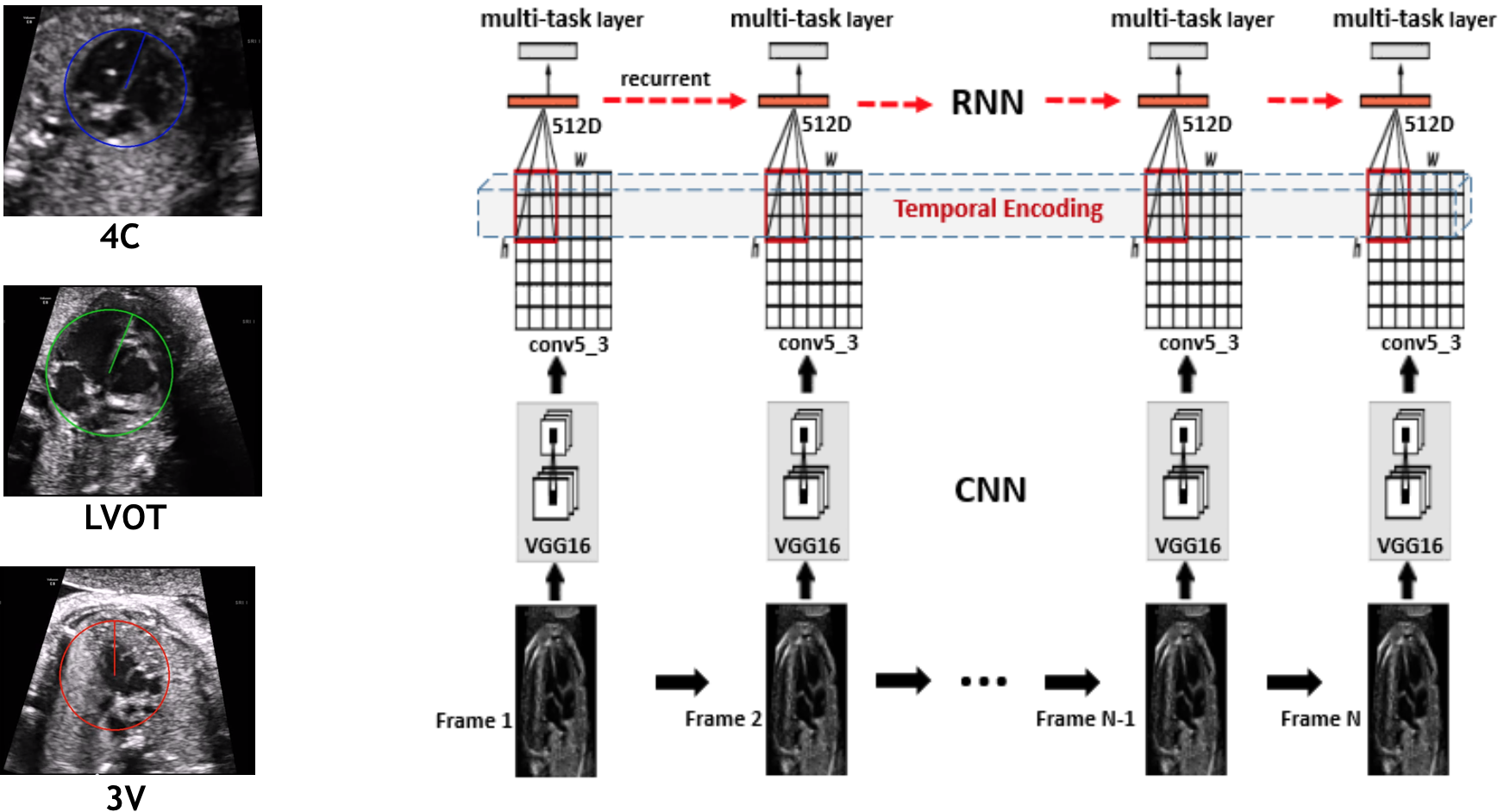}
\vspace*{-3mm}
\captionsetup{font=small}
\caption{\underline{Left}: three different view planes of fetal heart: the \textit{four chamber} (4C), the \textit{left ventricular outflow tract} (LVOT), and the \textit{three vessels} (3V). \underline{Right}: architecture of the proposed temporal heart network. }
\label{fig:main}
\end{figure}

Our goal is to jointly predict multiple key variables of the fetal heart at the frame level. These variables were originally defined in \cite{Bridge2017},  including the visibility, view plane, location and orientation of the fetal heart. We cast such a multi-task problem into joint classification and  regression tasks as follows. (i) The visibility indicates the present of the heart in the current frame. When the heart is present, the view plane can be categorized into one of three different views: the \textit{four chamber} (4C) view, the \textit{left ventricular outflow tract} (LVOT) view, and the \textit{three vessels} (3V) view, as shown in Fig. \ref{fig:main}. We consider them jointly as a 4-category classification problem, including a background class. (ii) The location of the heart is measured jointly by its center and radius, while the orientation is defined anti-clockwise from the increasing \textit{x}-direction, $\theta \in [0, 2\pi)$.  Therefore, both can be treated as a regression problem.


We formulate these problems as a multi-task learning model which can be jointly optimized by using  a joint classification, localization and orientation loss:
$L = L_{cls}+\lambda_1L_{loc}+\lambda_2L_{ori}$.
In our model, such a multi-task prediction is computed at each spatial location of the convolutional maps, by densely sliding a $3 \times 3$ window spatially through the last convolutional layer of the 16-layer VGGnet \cite{Simonyan2015}, as shown in Fig. \ref{fig:main}. Each sliding window generates a 512-D deep representation, which is computed using a fully-connected (fc) layer shared by all windows. The 512-D features are shared with three tasks by directly connecting to the output layer. \textit{This sliding window approach allows the model to focus on local details at each prediction, which is essential for accurate heart description, where the heart information is relatively weak in the whole US image}. Directly predicting the heart parameters from a whole image would reduce the accuracy considerably. Building on this multi-task framework, we develop a number of technical improvements that achieve considerable performance gains when combined.

\subsection{Circular Anchor Mechanism}
Anchor mechanisms have been applied in recent object detectors \cite{Ren2015}. We adopt this anchor mechanism within our heart analysis network. Unlike the rectangular anchors used in \cite{Ren2015}, we design circular anchors that work on the approximately circular appearance of the fetal heart in US images.  The function of an anchor is to provide a pre-defined reference that encourages the model to regress the parameters of the heart towards this reference. By introducing the circular anchors with various radii,  our model is able to detect the heart at multiple scales. 

A circular anchor is parameterized by a center and a radius. We design four such anchors, with radii of $\{80, 120, 160, 240\}$ pixels, at each spatial location on the convolutional maps. Specifically, the four anchors have the same centers as the associated $3 \times 3$ window, and share 512-D features. Each anchor has an independent group of multi-task results, including a 4-D softmax vector, and a 4-D  vector for the orientation and relative centre and radius of the heart. Therefore, for an input image, our model has $4wh$  anchors in total,  resulting in $4wh$ groups of predicted results, where the group with the highest classification score is selected as the final prediction. Here, $w$ and $h$ are the spatial  width and height of the convolutional maps. They are determined by the size of the input image which can be an arbitrary size by using a fully convolutional network \cite{Long2015}.


Training loss is computed at the anchor level. Each anchor has its own ground truth (GT), which can be pre-computed by using the anchor location and the GT region of the heart (as described in Sect. 3). The location of an anchor can be computed by mapping a spatial  location in the convolutional maps onto the input image. Anchor-level loss is computed in training, and the overall loss is a sum over all anchors  in a mini-batch. Following \cite{Ren2015}, we use a softmax classifier for the classification task, and apply a smooth-$l_1$ loss for the regression tasks.



\subsection{Intersection over Union (IoU) and Cosine Loss Functions}
IoU loss was recently applied for face detection \cite{Yu2016} and text detection \cite{Zhou2017}. 
A key advantage is that error is measured by an IoU of two rectangles, which considers multiple coordinate parameters as a whole unit. These parameters are strongly correlated,  but are optimized separately by using a smooth-$l_1$ or $l_2$ loss.

We adopt the IoU loss in our heart analysis network for the \textit{localization} task. Here we simply use a rectangle to represent the round shape of the heart, due to the easy differentiability of the rectangle IoU region, as shown in \cite{Yu2016}. At each spatial location in the convolutional layer, our model predicts a bounding box parameterized  by four values, $\textbf{x} = \{x_t, x_b, x_l, x_r\}$, which indicate the distances of the current point to the top, bottom, left and right sides of an associated bounding box. These distance values are calculated by projecting the spatial locations on the convolutional maps onto the input image, and the IoU loss is computed at each spatial location $(i, j)$, $L_{loc}^{(i,j)}=-\log ((\hat{A}\cap A)/(\hat{A} \cup A))$. $\hat{A}$ and $A$ are areas of  the GT and the predicted bounding boxes.  $\hat{A}\cap A$ and $\hat{A} \cup A$ can be computed from the 4-D parameters, $\textbf{x}$ and $\hat{\textbf{x}}$. Details are presented in \cite{Yu2016}, where the authors shown that such an IoU loss is differentiable, so that it can be applied for training a CNN with back-propagation. For the orientation task, we adopt a cosine loss introduced in \cite{Zhou2017}, $L_{ori}^{(i,j)}=1-\cos(\hat{\theta} - \theta)$.

\begin{figure}[tb]
\tiny
\includegraphics[height=0.91cm]{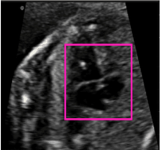} \qquad
\includegraphics[height=0.91cm]{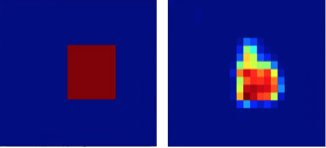} \qquad
\includegraphics[height=0.91cm]{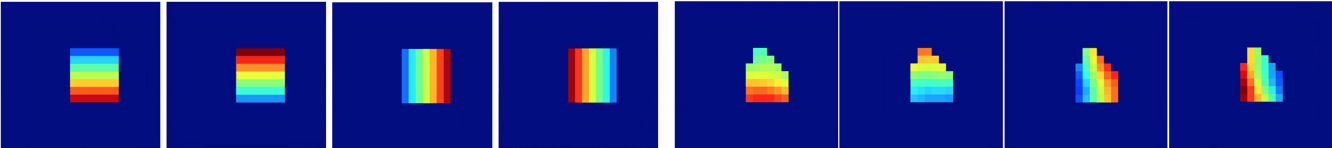}
\vspace*{-5mm}
\captionsetup{font=small}
\caption{GT and predicted maps for IoU layer.  \underline{Left}: rescaled input image. \underline{Middle}: the classification map with positive samples presented (GT v.s. predicted), other three classification maps (GT) are zeros. \underline{Right}: four localization maps (GT v.s. predicted). All GT maps are zeros if the heart is not present. The value is reduced from RED to Blue, and a non-blue region indicates positive locations where the IoU loss is computed.}
\label{fig:gt_maps_iou}
\end{figure}

We introduce a new IoU layer, which directly computes a multi-task prediction at each spatial location, without setting multiple anchors. This reduces the number of multi-task predictions from $\underline{4wh}$ to $\underline{wh}$, resulting in 9 prediction maps with size of $w \times h$: four maps for softmax classification, four maps for localisation and one map for orientation. A softmax map corresponding to the correct class and four localization maps are shown in Fig. \ref{fig:gt_maps_iou}. We choose  the location having the highest non-background classification score to output the final results.


In the training process, we generate 9 corresponding GT maps for each input image.  The GT bounding box of the heart is projected onto the last convolutional layer, and the positive samples are defined in the central area of the projected heart - a square region with sides of 0.7 times the diameter of the heart. Samples located outside the heart area are considered as negative ones. 
A classification map with positive samples and four localization maps are shown in Fig. \ref{fig:gt_maps_iou}.


\subsection{Temporally Recurrent Network}


To incorporate temporal information into our detection network, we design \textit{region-level} recurrent connections that compute temporal information at each spatial location of the convolutional layer, as shown in Fig. \ref{fig:main}. The original fc-layer computing  a 512-D feature vector from each sliding window is replaced by this recurrent layer, which is shared by all sliding windows through the convolutional maps. This allows each prediction to be conditioned on the same spatial location over sequential frames.  Therefore, our temporal model works at a local regional level of the frames, and can focus on local details of the heart. Intuitively, this makes it more discriminative and accurate for  describing heart details, which is a more principled design than recent recurrent models built on whole images \cite{Chen2015}.

We take a classical bi-directional LSTM \cite{Graves2005} for describing our recurrent architecture, where other types of RNN designs are readily applicable. It has two separate 256-D LSTM hidden layers that process the input sequence forward and backward, respectively. The inputs to each LSTM layer contain two components: $3 \times 3 \times 512$ convolutional features computed from current window at the current frame, and a 512-D temporal vector from the previous frame. This recurrent layer is connected to the output layers of the three tasks. By using a bi-directional RNN, our model is able to encode regional temporal context over the whole video, which is important to make a more reliable prediction. 

\section{Implementation Details}
\textbf{Anchor labels.} The labels of an anchor are defined by computing an IoU between a circular anchor and the heart GT. A positive anchor (one of three view planes) has an IoU  of $>0.7$; while a negative one is with an IoU of $<0.5$. Heart center and radius labels are computed by using  relative coordinates, as in \cite{Ren2015}. All values of a negative anchor are set to 0. For training, we select 256 anchors from each image which are unbalanced with a small number of the positive ones.


\textbf{Sequence generation.} It is difficult to train our temporal heart network by using  a limited number of fetal cardiac screening videos (e.g., only 91 videos in total). We generate a large number of video sequences by  splitting each video into a number of short random-length sequences (clips), including 25 - 50 continuous frames. In this way, we generate about 1,100 sequences, and this random sampling scheme is repeated at each epoch. Furthermore, our RNN model works on each sliding window though the convolutional layer, which naturally generate $w \times h$ times sequence samples. This increases the diversity of training samples, allowing our model to be trained efficiently, by avoiding  significant overfitting.  

\textbf{Training details.} We use the pre-trained 16-layer VGGnet \cite{Simonyan2015}. All new layers are initialized by using random weights. 
Weight values for different tasks are simply set to $\lambda_1=\lambda_2=1$. Our model with the anchor mechanism is trained by using a single-image mini-batch with a learning rate of 0.01, while the IoU-layer model has an 8-image mini-batch by using a learning rate of 10e-6. 
Both models are  trained for 500K iterations. Due to the limitation of GPU memory (12GB), end-to-end training of the Temporal HeartNet only allows for using a smaller-scale CNN, such as AlexNet \cite{Krizhevsky2012}. In current implementations with the VGGnet, the RNN component is trained separately for 50K iterations, by using a mini-batch of 4 sequences with a learning rate of 0.01.  


\section{Experiments, Results and Discussions}
\begin{figure}[tb]
\centering
\includegraphics[height=2cm]{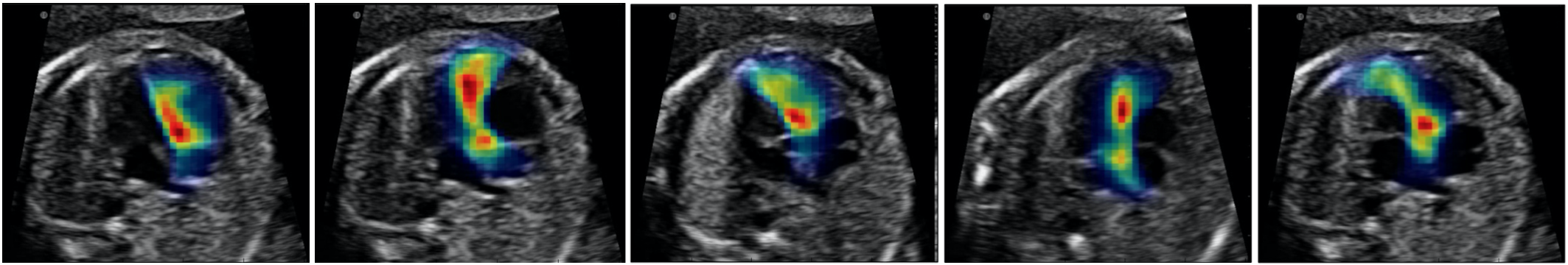}
\vspace*{-3mm}
\captionsetup{font=small}
\caption{Attention maps of the heart orientation on test images, which are computed by the Excitation Backprop scheme described in \cite{Zhang2016}.}
\label{fig:att_maps_iou}
\end{figure}

\textbf{Dataset.} The data was provided by the authors of \cite{Bridge2017}. It includes 91 fetal cardiac screening videos drawn from 12 subjects during routine clinical scans. The videos were annotated at the frame level, and each video has one or more of the three views of the fetal heart. We followed \cite{Bridge2017} by using leave-one-out cross-validations over each subject, resulting in 12-fold validations. \textbf{Evaluation protocols.}  Two protocols were used to evaluate the joint accuracy of classification and localization. (i) Following \cite{Bridge2017}, the heart was considered to be correctly detected if the heart view was correctly classified, and the predicted heart center was within $0.25 \times r$ of the annotated center. (ii) We consider a positive prediction if the heart view was correctly classified, and with an IoU over 0.25 with the GT location, which \textit{considers the prediction of heart radius}. Heart orientation error was measured as: $0.5 \times (1-\cos(\hat{\theta} - \theta))$, between the prediction and GT \cite{Bridge2017}. The errors were first averaged over all frames in a video,  then over all videos of each subject, and finally a mean value of 12-fold cross-validations was reported.\\
\\
\textbf{Experimental results.}  First, we investigate our multi-task learning framework where the deep features are shared by all tasks. We show that our model is capable of automatically learning task-specific features, allowing the shared features to focus on a particular region of attention for each single task. 
Fig. \ref{fig:att_maps_iou} shows the learned attention maps, which are computed by back-propagating a predicted orientation value to the input image, as described in \cite{Zhang2016}.  The map approximately indicates a key location that defines the orientation of the heart, suggesting that our model learns features discriminatively for each task, by only providing image-level supervisions. Second, our models were evaluated on real-world fetal cardiac screening videos. As shown in Table 1, the proposed temporal model  improves on its CNN counterpart (with circular anchor) under both evaluation protocols, reducing the errors as $28.8\%\rightarrow 21.6\%$ and $30.3\%\rightarrow 27.7\%$, respectively. The main improvement is obtained in classification. As expected, our model with an IoU layer achieves a higher localization accuracy over the circular anchor using a smooth-$l1$ loss. All three models achieve a high accuracy on orientation prediction, with marginal differences between them. \\

\begin{minipage}[b]{0.9\linewidth}

\begin{center}

\begin{tabular}{l|c|c|c|c|c}

Method & Protocol-I  & Protocol-II  & Class.  &Local. & Orient.  \\
\hline
\hline
Circular Anchor (CA) & 28.8 & 30.3  &21.9 &15.4 &0.074 \\
IoU loss layer & 26.8 &28.7 &22.0 &\underline{9.6}  & 0.084 \\
Temporal Model (RNN +CA)  & \underline{21.6} &\underline{27.7}  &\underline{16.1}  & 14.1& 0.072 \\

\hline
\end{tabular}

\end{center}
\captionsetup[table]{skip=-10pt}
\captionsetup{font=small}
\captionof{table}{Results on a real-world fetal cardiac screening video dataset (in error rates (\%)), including single classification and localization ($<0.25$ IoU).  }\label{tab:results} 
\end{minipage}

\textbf{Comparisons and discussions.} We compare our results with \textit{ Bridge et. al.}'s results (evaluated by Protocol-I), which are the state-of-the-art for this task \cite{Bridge2017}. Our IoU model achieves an error of 26.8\%, comparing favorably  against the best result (34\%) of \cite{Bridge2017}, without a temporal component.  This result is close to human inter-rater variation, which is about 26\% \cite{Bridge2017}. Our temporal model obtains an error of 21.6\%, showing that our automatic estimates are more consistent than human annotators. By using a CRF-filter, the temporal model of \cite{Bridge2017} obtains a low error of about 18\%, which is the best result evaluated by Protocol-I. However, it is important to note that the Protocol-I does not consider the prediction of heart radius, which is crucial for accurate heart detection.  The methods in \cite{Bridge2017} assume the heart radius is known, while our models include this prediction, allowing it to work in a more unconstrained environment.  Besides, the potential of our RNN may be not fully explored by training it on such limited video data. Better performance can be expected by using a more advanced training strategy.

\section{Conclusions}
We have presented new CNN-based approaches for automatically describing key parameters in fetal cardiac screening video. We formulated this problem as a multi-task heart prediction within detection framework, where a circular anchor mechanism or an IoU loss layer can be incorporated to improve the detection accuracies.
 A new recurrent architecture was designed and integrated seamlessly into our CNN, allowing it to encode \textit{region-level} temporal information through a video. These technical improvements result in a powerful temporal heart network, which achieves impressive results on a real-world clinical dataset.  
\\
\\
\textbf{Acknowledgments}. This work was supported by the EPSRC Programme Grant Seebibyte (EP/M013774/1).



{\small
\bibliographystyle{splncs03}
\bibliography{egbib2}}
\end{document}